\documentclass{article} % For LaTeX2e
\usepackage{collas2025_conference,times}
\usepackage{easyReview}
\makeatletter
\soulregister\cite7       % LaTeX’s \cite
\soulregister\citep7      % natbib \citep
\soulregister\citet7      % (optional) natbib \citet
\makeatother

\usepackage{longtable}
\usepackage{amsmath,amsfonts,bm}

\def\eqref#1{equation~\ref{#1}}
\def\1{\bm{1}}

\DeclareMathAlphabet{\mathsfit}{\encodingdefault}{\sfdefault}{m}{sl}
\SetMathAlphabet{\mathsfit}{bold}{\encodingdefault}{\sfdefault}{bx}{n}

\usepackage{ifthen}
\newboolean{addAck}

\usepackage{hyperref}
\hypersetup{
    colorlinks=true,
    linkcolor=red,
    filecolor=magenta,
    urlcolor=blue,
    citecolor=purple,
    pdftitle={Overleaf Example},
    pdfpagemode=FullScreen,
    }
\usepackage{times}
\usepackage{latexsym}
\usepackage{cuted}
\usepackage{color}

\usepackage{booktabs}
\usepackage{pifont}
\usepackage{multirow}

\usepackage[utf8]{inputenc}

\usepackage{float}
\usepackage{microtype}

\usepackage{inconsolata}

\usepackage{graphicx}
\usepackage{amsmath}

\usepackage{comment}
\usepackage{booktabs}
\usepackage{pifont} 
\usepackage{enumitem}
\usepackage{subcaption}
\usepackage{array}
\usepackage{wrapfig}

\usepackage{pifont}

\title{Improving Multimodal Large Language Models \\Using Continual Learning}

\author{
  \textbf{Shikhar Srivastava\textsuperscript{1}\thanks{Corresponding author: shikhar.srivastava@rochester.edu}}, \space
  \textbf{Md Yousuf Harun\textsuperscript{2}},
  \textbf{Robik Shrestha\textsuperscript{1}},
  \textbf{Christopher Kanan\textsuperscript{1}}
\\
\\
  \textsuperscript{1}University of Rochester,
  \textsuperscript{2}Rochester Institute of Technology
\\
}

\setboolean{addAck}{true} % Uncomment for camera-ready version, but NOT for submission.
\collasfinalcopy % Uncomment for camera-ready version, but NOT for submission.

\begin{document}
\maketitle
%\vspace{-2em}
\begin{abstract}
%\vspace{-0.5em}
Generative large language models (LLMs) exhibit impressive capabilities, which can be further augmented by integrating a pre-trained vision model into the original LLM to create a multimodal LLM (MLLM). However, this integration often significantly decreases performance on natural language understanding and generation tasks, compared to the original LLM. This study investigates this issue using the LLaVA MLLM, treating the integration as a continual learning problem. We evaluate five continual learning methods to mitigate forgetting and identify a technique that enhances visual understanding while minimizing linguistic performance loss. Our approach reduces linguistic performance degradation by up to 15\% over the LLaVA recipe, while maintaining high multimodal accuracy. We also demonstrate the robustness of our method through continual learning on a sequence of vision-language tasks, effectively preserving linguistic skills while acquiring new multimodal capabilities. Project webpage: \url{https://shikhar-srivastava.github.io/cl-for-improving-mllms}
\end{abstract}

%%%%%%%%%%%%%%%%%%%%%
%\vspace{-1.5em}
\section{Introduction}

Advances in integrating visual information with large language models (LLMs) have led to the development of multimodal large language models (MLLMs), excelling at many vision-language (VL) tasks~\citep{wu2023visual, yang2023mm, li2023blip, alayrac2022flamingo, vicuna2023, driess2023palm, zhu2023minigpt, liu2023llava, luo2023cheap, koh2023grounding, lin2023vila,liu2023improved}. Recent studies converge on a general recipe for developing MLLMs: Alignment of LLM token embeddings with visual embeddings followed by instruction-tuning on VL tasks like visual question answering (VQA)~\citep{liu2023llava}. {However, creating an MLLM often degrades the LLM's natural language understanding (NLU) and generation (NLG) performance~\citep{li2023blip,driess2023palm}. This phenomenon of the sudden degradation in previously learned capabilities as a result of progressive training is known as catastrophic forgetting~\citep{mccloskey1989catastrophic}}. For instance, {the multimodal} PaLM-E experienced an 87\% {\textit{relative} drop in performance on a range of NLG tasks compared to} its unimodal base LLM~\citep{driess2023palm}. {Recent approaches towards MLLMs, such as LLaVA, notably also suffer from the loss of linguistic abilities~\citep{liu2023llava,lin2024vila}. This degradation occurs primarily due to the significant distribution shift when adapting LLMs, originally trained on text-only tasks, to multimodal vision-language (VL) contexts. Limiting this degradation by freezing the LLM, training secondary modules~\citep{dai2024instructblip,koh2023grounding,zhu2023minigpt} underperforms fully adapted approaches like LLaVA and precludes emergent multimodal capabilities like in-context learning ~\citep{yu2023mm,lin2024vila}. Linguistic forgetting, therefore, is pervasive in the multimodal adaptation of LLMs, with limited work on studying this phenomenon and developing methods to efficiently mitigate it.}

{We posit that a practical MLLM must excel in both purely text-based \textit{and} vision-language tasks. In real-world scenarios, users frequently alternate between text-only and multimodal queries in multi-turn interactions -- interacting with multimodal systems by asking questions in text, then refining it with an image, or vice versa~\citep{chen2024tomgpt,chen2024we}. Moreover, strong linguistic understanding remains essential for performative responses even inside multimodal prompts~\citep{lin2023vila}. 
To address these challenges, we propose employing Continual Learning (CL) approaches, explicitly developed to mitigate catastrophic forgetting arising from shifting task distributions~\citep{parisi2019continual}. Because the introduction of visual information constitutes a significant distributional shift, studying MLLM adaptation within a CL framework is natural, and facilitates preservation of linguistic capabilities.}

{Finally, a general multimodal learning system must continuously acquire new multimodal capabilities without sacrificing both pre-existing linguistic and acquired multimodal capabilities. In CL, a sequence of non-stationary tasks is learned continuously. We treat the first task as the unimodal pre-training task already learned by the base LLM, followed by new VL tasks. We study two paradigms. In the first, we seek to recreate LLaVA 1.5 while mitigating linguistic forgetting through our methods. In the second, we sequentially learn each VL dataset in the LLaVA recipe, evaluating the model’s incremental acquisition of multimodal skills without degrading previously learned capabilities.}

\begin{figure}[t]
    \centering
    \includegraphics[width=0.75\linewidth]{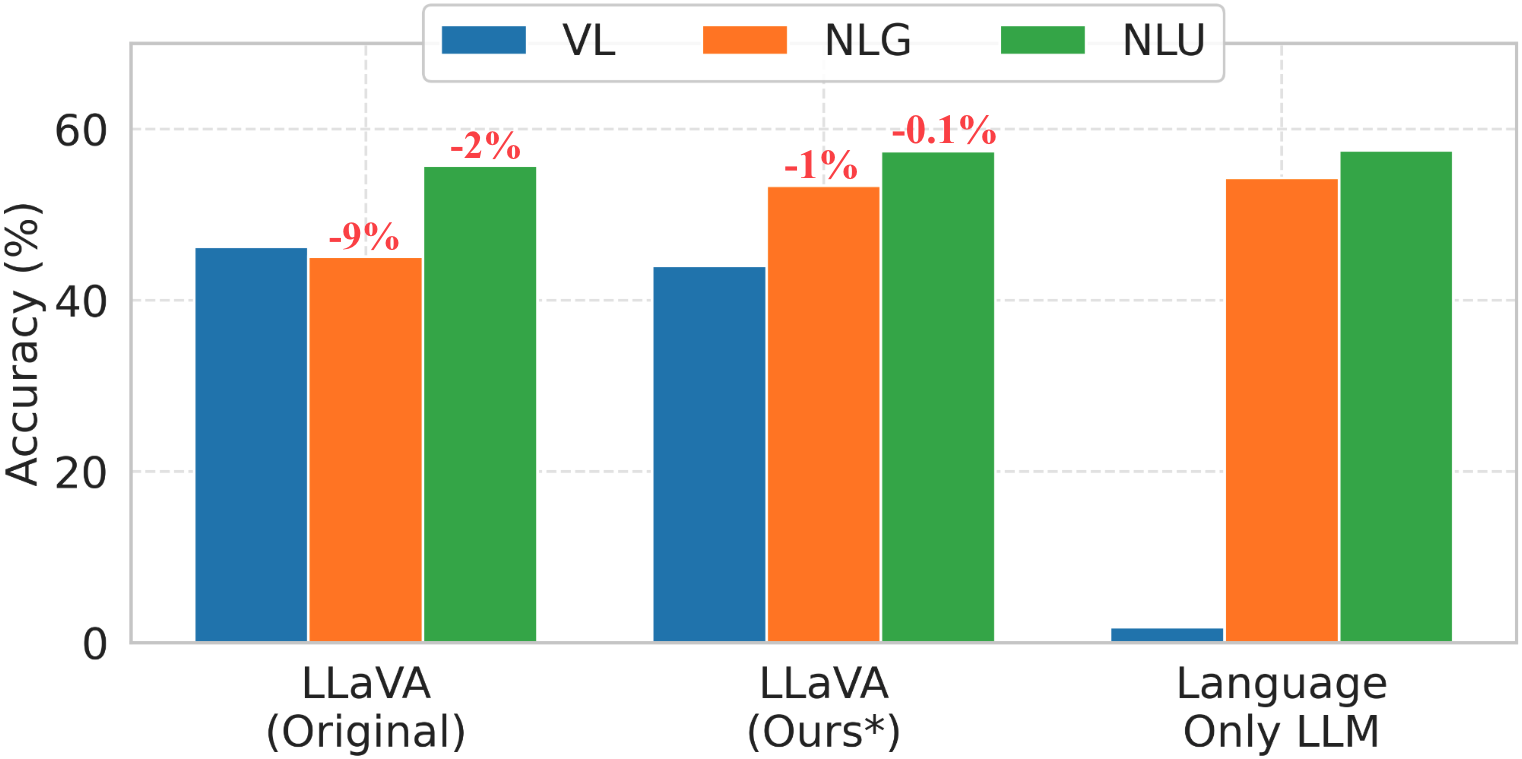}
    \caption{
    Summary results of the best continual learning method we evaluated for training LLaVA 1.5 compared to the Language-Only LLM, and the original version of LLaVA 1.5. All results are with Pythia 2.8B as the base LLM. The best method has almost the same vision-language (VL) accuracy while providing a large increase in linguistic performance on 1 NLG and 4 NLU tasks by 8\% and 2\% (absolute), respectively.
    Bar annotations indicate linguistic forgetting relative to the Language-Only LLM, highlighted in \textcolor{red}{red}.
    }
    \label{fig:visual-abstract}
\end{figure}

%\clearpage

\paragraph{This paper makes the following contributions:}
\begin{enumerate}%[noitemsep,nolistsep]
    \item Using the original LLaVA 1.5 training recipe, we study linguistic forgetting in 9 MLLMs, including 5 built on the Pythia family of LLMs to study the role of model scales and instruction tuning on such linguistic forgetting.

    \item We study the effectiveness of 5 mitigation techniques for reducing linguistic forgetting and show that the best method improves accuracy for NLG, NLU, and VL tasks compared to the naive LLaVA recipe (see Fig.~\ref{fig:visual-abstract}).
    
    \item We pioneer studying CL for MLLMs by sequentially learning VL tasks, where we assess the efficacy of CL techniques to mitigate catastrophic forgetting in this challenging scenario. 
    
\end{enumerate}

%%%%%%%%%%%%%%%%%%%%%%%%
\section{Background}

%%%%%%%%%%%%%%%%%%%%%%%%
\subsection{Multimodal Large Language Models}

%% edited text
Open source MLLMs integrate pre-trained vision encoders into LLMs (see ~\cite{bordes2024introduction} for review).

Notable works include PaLM-E~\citep{driess2023palm}, LLaVA 1.5~\citep{liu2023llava,liu2023improved}, MiniGPT4~\citep{zhu2023minigpt}, LaVIN~\citep{luo2023cheap}, Visual ChatGPT~\citep{wu2023visual}, BLIP-2~\citep{li2023blip}, Qwen-VL~\citep{bai2023qwen},
Flamingo~\citep{alayrac2022flamingo}, FROMAGe~\citep{koh2023grounding}, and VILA~\citep{lin2024vila}. PaLM-E is an embodied MLLM that is capable of grounded reasoning for VL and language tasks. MiniGPT-4 uses a trainable fusion layer to link the LLM to a vision encoder, with both the LLM and vision encoder frozen. 
Similar to MiniGPT-4, Qwen-VL and Qwen-VL-Chat models use a one-layer cross-attention module to inject vision embeddings into an LLM. 
BLIP-2 uses a Q-Former to map visual information into the input space of LLM. LaVIN introduces Mixture-of-Modality Adaptation to extend the multimodal capabilities of LLMs with a vision encoder. FROMAGe demonstrated good performance on image captioning and image-text retrieval tasks using trainable input and output linear layers, while keeping the vision encoder and LLM frozen. In this paper, we study LLaVA, which shares many similarities with these models and is described in detail in Sec.~\ref{sec:llava}.

While many MLLMs have been proposed, most papers focus only on VL performance and do not measure whether this resulted in any loss of NLU/NLG performance, including the original MiniGPT-4 and LLaVA papers. Notable exceptions are PaLM-E, LaVIN, and FROMAGe. However, they do not study methods for mitigating forgetting when an LLM is converted into an MLLM.
{~\citet{driess2023palm} observed that multimodal adaptation led to \textit{catastrophic performance degradation} (\textit{over 87\%}) on a suite of 21 general language benchmarks, a problem especially pronounced in smaller MLLMs ($<$14B parameters). This is particularly concerning as smaller LLMs comprise the largest segment of open-source LLMs~\citep{gunasekar2023textbooks,li2023textbooks}, and are also the most widespread for consumer applications (such as on-device and edge deployments). Addressing this problem of linguistic forgetting could therefore significantly improve the utility of MLLMs. Approaches such as InstructBLIP~\citep{dai2024instructblip}, FROMAGe~\citep{koh2023grounding} keep the LLM and visual encoders frozen while learning secondary instruction-aware modules. These approaches limit interference in vision-language and linguistic capabilities, but have been shown to underperform LLaVA and preclude emergent multimodal capabilities like in-context learning ~\citep{yu2023mm,chen2024visual,lin2024vila}}.
The only work that has attempted to address NLU/NLG forgetting in a comparatively performative MLLM is  VILA~\citep{lin2024vila}, which did so by including MMC4 in their training recipe, which is a dataset that interleaves text and images~\citep{zhu2023multimodal}. Training for this interleaved image-text task also required a large increase in pre-training compute. Moreover, they only study its impact on a single NLU dataset, neglecting its potential impact on NLG tasks. Our work is the first to study mitigating linguistic forgetting in MLLMs using CL methods.

%%%%%%%%%%%%%%%%%%%%%%%%
\subsection{Continual Learning}

Creating an MLLM maps directly onto a CL scenario involving learning two tasks sequentially, where the first task is acquiring NLU/NLG capabilities and the second is acquiring VL capabilities. Although some recent works have studied CL with unimodal LLMs~\citep{luo2023empirical,lin2023speciality}, our work is the first to treat creating an MLLM as a CL problem. While there are many CL techniques, two general families of methods are most appropriate for MLLMs: rehearsal and regularization. Rehearsal methods store or reconstruct previously observed data that is mixed into new data to overcome catastrophic forgetting~\citep{hayes2021replay, hayes2020REMIND, chaudhryER_2019, harun2023siesta, hou2019learning, rebuffi2017icarl}. The primary challenge with using it for MLLMs is that the training sets for many open-source LLMs have not been released, and even if they were released, determining what to rehearse may be intractable due to the large size of these datasets. Regularization methods adjust training processes and impose constraints on weight updates to reduce forgetting~\citep{chaudhry2018efficient, chaudhry2018riemannian, dhar2019learning, kirkpatrick2017overcoming, zenke2017continual}. We focus on lightweight regularization methods in this work that can scale to MLLMs.

%%%%%%%%%%%%%%%%%%%%%%%%

Parameter Efficient Tuning (PET) methods are an efficient alternative to full fine-tuning~\citep{houlsby2019parameter}. They update fewer parameters, thereby reducing computational costs while maintaining performance. In \cite{harun2024overcoming}, LoRA was shown to effectively retain performance in ImageNet pre-trained models as new classes were continually learned. Inspired by this, we study the effectiveness of LoRA~\citep{hu2021lora} in mitigating catastrophic forgetting while learning VL tasks.

%%%%%%%%%%%%%%%%%%%%%%%%%%
\section{Methods}

%%%%%%%%%%%%%%%%%%%%%%%%%%
\subsection{The LLaVA MLLM}
\label{sec:llava}
We study LLaVA 1.5, henceforth referred to as LLaVA, which is one of the most widely used multimodal training protocols. LLaVA has the following components:
\begin{itemize}[noitemsep,nolistsep]

    \item \textbf{Visual Encoder:} Following earlier implementations~\citep{liu2024visual,liu2023improved}, we use a pre-trained ViT-L/14 from CLIP which takes an image resolution of 336px as the vision encoder, which is kept frozen throughout the training to ensure stability and prevent overfitting to the initial training tasks.
    \item \textbf{LLM:} We study 9 LLMs of various sizes and training protocols (see Sec.~\ref{sec:llms}).
    \item \textbf{Alignment Network:} To inject other modalities into LLaVA, it uses a linear alignment network that projects embeddings from vision into the representational space of the text tokens~\citep{merullo2022linearly}. This network has two layers that map each vision embedding to the text embedding space. 
\end{itemize}
We follow the standard LLaVA 1.5 training recipe, with details, including optimizer settings and hardware, provided in Section \ref{sec:llava-training-details} and Appendix ~\ref{sec:hardware-and-compute}.

%%%%%%%%%%%%%%%%%%%%%%%%%%%%%%%%%%%
\subsubsection{LLMs Studied}
\label{sec:llms}
We study 9 choices for LLaVA's LLM: 6 Pythia models~\citep{biderman2023pythia}, Phi2 (3B)~\citep{gunasekar2023textbooks}, and 2 LLaMA 2 (7B) models. The 6 Pythia models are of various scales (160M, 410M, 1B, 1.4B, and 2.8B parameters), where there are 2 versions of Pythia (1.4B), one of which is instruction fine-tuned. Likewise, we study the original version of LLaMA 2 and an instruction fine-tuned version, Vicuna-1.5 7B~\citep{vicuna2023}, which was used in LLaVA 1.5.

Phi2~\citep{li2023textbooks} has been trained on the same dataset as Phi1~\citep{gunasekar2023textbooks}, which includes a curated selection of ``textbook quality'' data from the web (6 billion tokens) and an additional 1 billion tokens of synthetically generated textbooks and exercises created using GPT-3.5~\citep{brown2020language}. The Phi series {contain performant models given their small parameter size class such as Phi2 with 2.7B parameters}~\citep{li2023textbooks}.
Pythia~\citep{biderman2023pythia} comprises two sets of 8 models, each corresponding to two datasets. For every model size, one set is trained on the Pile dataset~\citep{gao2020pile}, while the other set is trained on a version of the Pile where global de-duplication has been applied. The granular model scaling suite of Pythia is particularly useful for our study. We select the de-duplicated set of Pythia models. LLaMA 2 is reportedly trained on a mix of publicly available online data, but specific details are not available~\citep{touvron2023llama2}. LLaVA 1.5 reports its best performance with instruction fine-tuned LLaMA 2 LLMs~\citep{liu2023improved}.

%%%%%%%%%%%%%%%%%%%%%%%%%%
\subsection{Continual Learning Methods}

To mitigate catastrophic forgetting in MLLMs, we examine several methods:
\begin{enumerate}
    \item \textbf{Naive Fine-Tuning} corresponds to the original LLaVA method with no modifications. 
    
    \item \textbf{LoRA} keeps the original LLM weights frozen and learns low-rank updates for them~\citep{hu2021lora}. Following LoRA's recommended protocol, we inject LoRA weights into all LLM linear layers. After each task is learned, the LoRA weights are merged into the LLM. Details are given in Appendix~\ref{sec:lora-hparam}.

    \item \textbf{Soft Targets}, introduced in \citet{harun2024overcoming}, aim to mitigate forgetting in CL. However, they are designed for image classification tasks and are not directly applicable to MLLMs. We adapt soft targets to MLLMs by dynamically re-parameterizing the conditional token distribution. 
    We update the hard target vector $Y$ by reducing the target tokens by \(- \alpha\), and offsetting non-target tokens by \(+ \alpha/(N - 1)\), where $\alpha$ controls the smoothing, and \(N\) is the LLM's vocabulary size. We discuss the choice of \(\alpha\) in Appendix~\ref{sec:soft-target-hparam}.
 
    \item \textbf{Rehearsal (Experience Replay)} is an effective method for CL that involves storing data from earlier tasks and mixing it with data from current task. We study it in our CL experiments since {in practice, the pretraining data for the base LLMs (i.e. task 1) are inaccessible}. We study storing 1\% of randomly selected samples from each previous task, excluding task 1.
    
    \item \textbf{mSGM} is based on SGM, which combines soft targets, weight initialization, and LoRA to mitigate catastrophic forgetting~\citep{harun2024overcoming}. We use a version that omits the weight initialization and output layer freezing process used by SGM for class incremental learning since the output vocabulary is static and output layers are used for the causal generation of LLMs. \cite{harun2024overcoming} found that SGM was especially effective when paired with rehearsal.

\end{enumerate}
In all methods, the alignment layer and LLM are trained and the ViT is frozen. Following the LLaVA 1.5 training protocol, the VL datasets are trained in a single epoch corresponding to a single training pass through each dataset after learning task 1 to update the LLM.

%%%%%%%%%%%%%%%%%%%%
%\vspace{-0.5em}
\subsection{Evaluation Datasets}
\label{sec:datasets}
To facilitate analysis across tasks, all of our evaluation sets use the same metric: accuracy.

\textbf{Natural Language Evaluation:}
We use six datasets. For NLG, we use Lambada~\citep{paperno2016lambada}, which is the only NLG dataset that uses accuracy for evaluation. For NLU, we use ARC-Easy~\citep{clark2018think}, ARC-Challenge~\citep{clark2018think}, Winogrande~\citep{sakaguchi2021winogrande}, and WSC~\citep{sakaguchi2021winogrande} for NLU. For completeness, we also study multiple NLG datasets using a `\% Change' \(\Delta'\) metric in Appendix~\ref{sec:multiple_nlg_analysis}.

\textbf{Vision-Language Evaluation:} We use the test sets corresponding to each VL dataset used for training LLaVA: VQAv2 and GQA for general VQA tasks~\citep{goyal2017making, hudson2019gqa}, TextVQA OCR (and Pure) for OCR tasks~\citep{singh2019towards}, and RefCOCO for referential expression generation tasks~\citep{kazemzadeh2014referitgame}. 
We use the \emph{slim} version of all test datasets to make our experiments tractable~\footnote{\url{https://github.com/TRI-ML/vlm-evaluation}}. Additional details are given in Appendix~\ref{sec:dataset_prep}.

%\vspace{-0.5em}
\subsection{LLaVA Training Details}
\label{sec:llava-training-details}\label{app:exp_settings}
Following the LLaVA 1.5 protocol ~\citep{liu2023llava}, the visual encoder is a CLIP VIT-L@336px with an image resolution of 336px with letterbox resizing, and the alignment network is a two-hidden layer MLP projector with GELU activation. The LLM generations are limited to a maximum token length of \(2048\). For training, we used the Adam optimizer, with FSDP (Fully Sharded Data Parallel), gradient checkpointing and mixed precision to train all our models~\citep{zhao2023pytorch}. 

\textbf{Alignment Stage:} During LLaVA 1.5's alignment stage the LLM is kept frozen, while the alignment network is plastic. We use a learning rate of \(0.001\) with a linear warmup followed by a cosine decay scheduler. The training process is carried out for \(1\) epoch with a global batch size of \(256\), distributed as \(16\) samples per device. During the alignment stage, we shard only the gradients and optimizer states, not the parameters. 

%\vspace{-2em}
\textbf{Fine-tuning Stage:} During LLaVA 1.5's fine-tuning stage both the LLM and alignment networks are jointly fine-tuned. We set the learning rate to \(2e-05\), with a linear warmup followed by cosine decay. This stage also runs for 1 epoch but with a smaller global batch size of \(128\), keeping the per-device batch size constant at \(16\). Depending on the size of the LLM (Phi2 3B vs Pythia 160M), we vary the per-device batch size, but the global batch size is kept constant across all experiments.
We switch to the FSDP full sharding strategy, with all parameters, optimizer states, and gradients sharded across the devices.

\textbf{Continual LLaVA:} For the Continual LLaVA setting, we follow the same experimental configuration as before in the LLaVA setting. A model trained on Task \((i)\), will then be simply used as the pretrained checkpoint for Task \((i+1)\), with training following normally. The training configurations are then identical to the fine-tuning stage. In the case of PET methods like LoRA, the adapter weights are merged back into the LLM before starting training on the new task. 

To reproduce the LLaVA 1.5 experiment, we build on top of the Prismatic library~\citep{karamcheti2024prismatic}, and use the Prismatic library for vision-language tasks and EleutherAI's LM\_Eval library~\citep{eval-harness} for natural language evaluations. All our code is written in PyTorch~\citep{paszke2019pytorch}.

%%%%%%%%%%%%%%%%%%%%%%%%%%%%
%\vspace{-0.5em}
\subsection{Measuring Performance}
%\shikhar{\% Change for NLU/NLG forgetting.}
To assess performance, we compute the forgetting metric $\Delta$, where a positive value indicates forgetting, and a negative value indicates learning the current task enhances the performance of previously acquired tasks (or a backward transfer~\citep{zenke2017continual}). The performance change, $\Delta$, on task $t$ after training on task $k$ is defined as:
\begin{equation}
    \Delta_t(k) = \omega_t(1) - \omega_t(k), \quad \forall t < T
\end{equation}
where $T$ is the total number of tasks, and {$\omega_t(k)$ is the harmonic mean of the accuracy values of the evaluation datasets for task $t$, after training on task \(k\)}. We use the harmonic mean because it is dominated by the accuracy of the \emph{worst-performing} dataset for the task, thereby emphasizing the need to perform well, avoid forgetting for all datasets. 
%%%%%%%%%%%%%%%%%%%%%%%%%%%%%%%%
%\vspace{-0.5em}
\section{Experiments}
%\vspace{-0.5em}

In our experiments, we treat training a MLLM as a CL problem, where the system learns a sequence of tasks from 1 to $T$. Task 1 always consists of training the LLM, where we assume the LLM has already been trained and we do not know the provenance of the training data or the exact methods to create it, which is true of many commonly used LLMs (e.g., Llama 2 and Llama 3). 
In our experiments on analyzing and mitigating linguistic forgetting, there are 2 tasks: 1) learning the base LLM, and 2) learning the mixture of VL datasets. In CL experiments, there are 5 tasks where the first is training the base LLM, and then each VL task is sequentially learned. 

Following the standard LLaVA recipe, for all experiments, task 2 begins by training the VL alignment network using the LLaVA-595 CC-SBU captioning dataset. The network is trained to generate captions with an auto-regressive loss, with the vision encoder and LLM kept frozen. Because only the alignment network is trained, the model has a modicum of visual ability without impairing its language capabilities. Subsequently, the LLM and alignment layer are trained for tasks 2 to $T$. For Secs.~\ref{sec:analyzing-linguistic-forgetting} and \ref{sec:mitigation-analysis}, their two-task setup is identical to LLaVA 1.5's training protocol. Additional implementation details are given in {Sec.}~\ref{app:exp_settings}.

%%%%%%%%%%%%%%%%%%%%%%%%%
\subsection{Analyzing Linguistic Forgetting}
\label{sec:analyzing-linguistic-forgetting}
Linguistic forgetting has not been studied in LLaVA, so before assessing CL methods, we first measure linguistic forgetting and VL performance for all of our LLMs using the standard LLaVA training recipe; which uses naive fine-tuning. Using this recipe, our 9 LLMs are transformed into MLLMs.
%\vspace{-0.5em}
\begin{figure*}[th!]
    \vspace{-4em}
    \centering
    \includegraphics[width=0.75\linewidth]{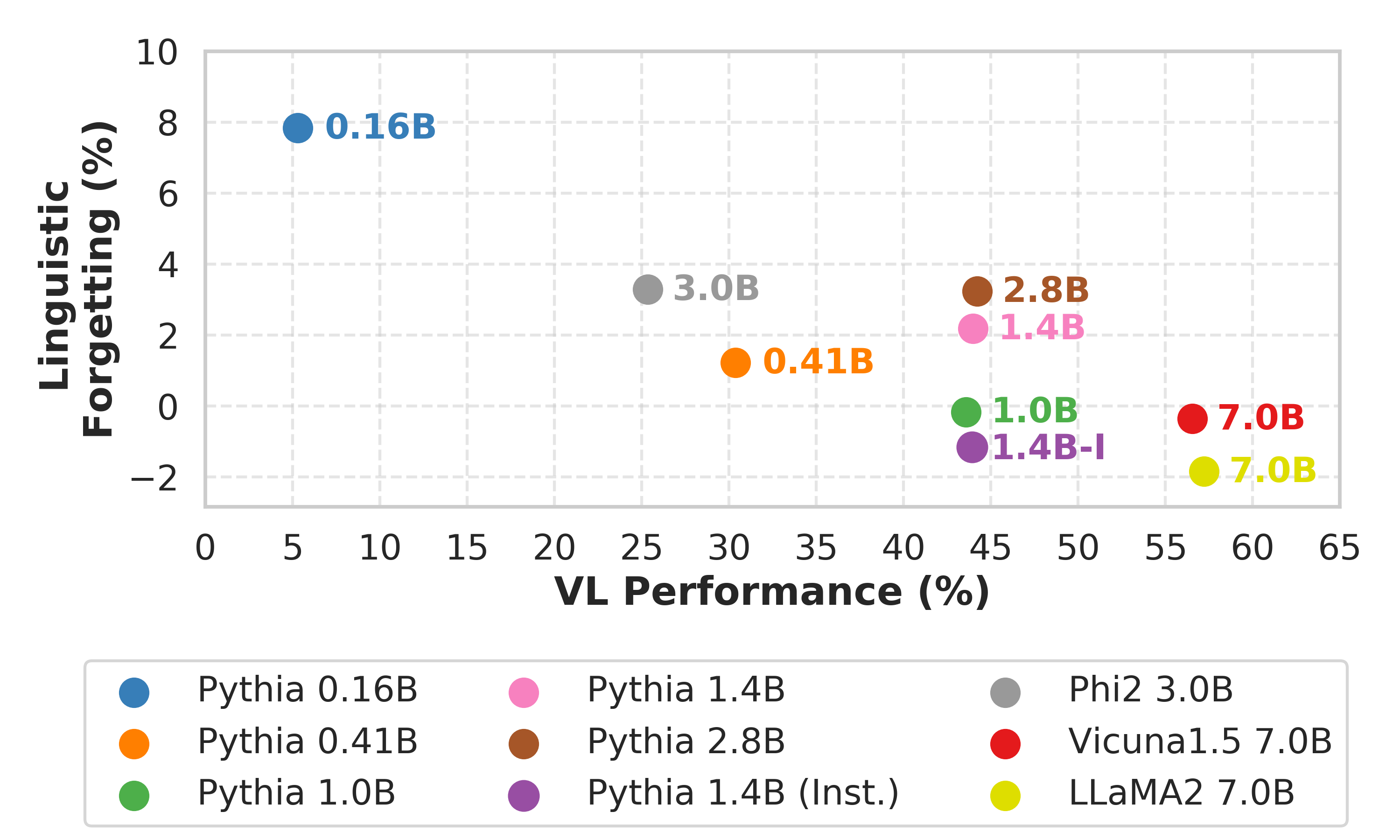}
    \vspace{-0.5em}
    \caption{Linguistic forgetting vs. VL performance for 9 MLLMs trained with the LLaVA recipe. Five models exhibited linguistic forgetting, and four had negative linguistic forgetting values, indicating that VL training resulted in positive transfer to language tasks.}
    \label{fig:forgetting-fig}
    \vspace{-1.5em}
\end{figure*}

\paragraph{Overall Results.} Our overall results are given in Fig.~\ref{fig:forgetting-fig}. Five models suffer from linguistic forgetting, while surprisingly, four have increased NLU/NLG accuracy due to positive transfer from the VL tasks. We next dive into these results, where our analysis of NLU/NLG forgetting as a function of model scale is provided in a later subsection. 

%\vspace{-0.5em}
%%%%%%%%%%%%%%%%%%%%%%%%%%%%%%%%
\paragraph{Positive Backwards Transfer \& Analysis of NLU vs. NLG Tasks.}
In Table~\ref{tab:nlu_vs_nlg}, we study the impact of forgetting on NLU vs. NLG tasks. We find that forgetting is much more severe for our one NLG task than the NLU tasks. 
All MLLMs exhibit greater NLG forgetting than NLU forgetting as shown in Table~\ref{tab:nlu_vs_nlg}. In terms of model size, typically smaller models show higher NLG forgetting, which is consistent with PaLM-E's observations. Across MLLMs, NLG forgetting is much more severe compared to NLU, and the NLU datasets are the source of the cases of positive backward transfer (negative \(\Delta\)). We posit this may be due to the additional common-sense reasoning and world knowledge encoded in visual-language tasks and instructions, which is relevant for the NLU tasks.  We include additional results on the NLU and NLG forgetting in Appendix~\ref{sec:analysis-nlu-nlg-forgetting} (see Fig.~\ref{fig:forgetting-nlu-nlg}), and analyze linguistic generation capabilities across multiple NLG benchmarks in Appendix~\ref{sec:multiple_nlg_analysis}.
%\vspace{-4em}

\begin{wraptable}{r}{0.52\textwidth}
    \vspace{-3.6em}
    \centering
    
    %--- First table ---
    \resizebox{0.9\linewidth}{!}{%
      \begin{tabular}{lrrrr}
        \toprule
        \textbf{LLM} & \textbf{Scale (B)} & \textbf{NLU $\Delta \downarrow$} & \textbf{NLG $\Delta \downarrow$} \\
        \midrule
        \multirow{6}{*}{Pythia} 
        & 0.16 & 0.94 & 12.01 \\
        & 0.41 & --1.19 & 8.62 \\
        & 1.0 & --1.63 & 4.95 \\
        & 1.4 & 0.55 & 8.07 \\
        & 2.8 & 1.74 & 9.18 \\
        \midrule
        Pythia (I) & 1.4 & --1.20 & --1.01 \\
        \midrule
        Phi2 & 3.0 & 2.60 & 4.39 \\
        \midrule
        Vicuna 1.5 & 7.0 & --0.98 & 2.04 \\
        \midrule
        LLaMA 2 & 7.0 & --2.15 & --0.43 \\
        \midrule
        Average & --- & --0.15 & 5.31 \\
        \bottomrule
      \end{tabular}%
    }
    \caption{\textbf{NLU vs NLG Forgetting:} Negative $\Delta$ indicates positive backward-transfer (desirable). The letter (I) denotes an instruction-tuned model.}
    %\vspace{-1em}
    \label{tab:nlu_vs_nlg}
    \resizebox{0.9\linewidth}{!}{%
      \begin{tabular}{l|c|c|cc}
        \toprule
        \textbf{Base LLM} & \textbf{Instr.} & \multicolumn{1}{c|}{\textbf{VL Avg.}} & \multicolumn{2}{c}{\textbf{NL Avg.}} \\
        & & \textbf{Acc $\uparrow$} & \textbf{Acc $\uparrow$} & \textbf{$\Delta \downarrow$} \\
        \midrule
        \multirow{2}{*}{LLAMA 2 (7B)} & \ding{55} & \textbf{57.22} & \textbf{66.23} & \textbf{~--1.84} \\
                                     & \ding{51} & 56.55 & 64.44 & ~--0.36 \\
        \midrule
        \multirow{2}{*}{Pythia (1.4B)} & \ding{55} & \textbf{43.97} & \textbf{45.51} & ~~~2.18 \\
                                       & \ding{51} & 43.93 & 41.37 & \textbf{~--1.16} \\
        \bottomrule
      \end{tabular}
    }
    \caption{Results for LLaMA 2 and Pythia after LLaVA training, with and without instruction fine-tuning. Negative $\Delta$ indicates positive backward transfer.}
    \label{tab:instruction-ft}
    \vspace{-3em}
\end{wraptable}

\paragraph{The Impact of Instruction Fine-Tuning.}
Previous work on continual learning in large language models (LLMs) has shown that instruction fine-tuning mitigates catastrophic forgetting~\citep{luo2023empirical}. This study extends the investigation to multimodal language models (MLLMs), analyzing LLaVA models trained with Pythia (1.4B) and LLaMA 2 (7B), and their instruction fine-tuned versions: Vicuna 1.5, used in the original LLaVA paper~\citep{liu2023improved}, and an instruction-tuned Pythia (1.4B)\footnote{\url{https://huggingface.co/lambdalabs/pythia-1.4b-deduped-synthetic-instruct}}. As shown in Table~\ref{tab:instruction-ft}, non-instruction-tuned LLMs exhibit minor increases in visual-linguistic (VL) accuracy ($<1$\%) and a larger difference in natural language understanding/generation (NLU/NLG) accuracy compared to their instruction-tuned counterparts. {The individual task accuracy breakups are shown in Tables~\ref{tab:instruct_tuned_nl_accuracies}, ~\ref{tab:instruct_tuned_vl_accuracies} in the Appendix}. Instruction-tuned LLaMA 2 (7B) and Pythia (1.4B) models show minimal or even negative linguistic forgetting, indicating positive backward transfer. Conversely, base LLMs display greater variability, with base Pythia experiencing more forgetting than base LLaMA 2, {possibly due to differences in their pre-training datasets affecting post-multimodal training linguistic performance}. Instruction tuning significantly reduces NLG forgetting in Pythia (1.4B) and induces positive backward transfer from VL to NLG, with a similar trend observed in NLU tasks. This suggests that instruction-tuned LLMs exhibit little to no forgetting, likely due to the distributional similarity between text-only instruction tuning tasks and visual-instruction tuning, helping to mitigate catastrophic forgetting.

%\pagebreak[7]
%%%%%%%%%%%%%%%%%%%%%%%%%
\subsection{Mitigating Linguistic Forgetting}
\label{sec:mitigation-analysis}

Following the results in Sec.~\ref{sec:analyzing-linguistic-forgetting}, we study the efficacy of our mitigation methods toward reducing linguistic forgetting.
Due to limited computational resources, we only exhaustively tested the mitigation methods with Pythia (160M), which was selected due to its small size. We then evaluate the best method using all Pythia models in Sec.~\ref{sec:model-scaling-para}. 

Our results with various mitigation methods on LLaVA training of Pythia (160M) are given in Table~\ref{tab:mitigation-results-pythia160m}. Soft Targets has the highest accuracy across VL datasets with the least linguistic forgetting. LoRA and mSGM better preserve NLU/NLG performance but at the cost of decreased VL accuracy compared to naive fine-tuning. 

\begin{table*}[!h]
  \centering
  \resizebox{0.9\linewidth}{!}{
    \begin{tabular}{l|cccc|c|cc}
     \toprule
     \textbf{Model} & \multicolumn{4}{c|}{\textbf{Vision-Language (VL) $\uparrow$}} & \textbf{VL Avg.} & \multicolumn{2}{c}{\textbf{NL Avg.}} \\
      & \textbf{VQAv2} & \textbf{TextVQA OCR} & \textbf{TextVQA Pure} & \textbf{GQA} & Acc $\uparrow$ & $\Delta \downarrow$ & Acc $\uparrow$ \\
     \midrule
Pythia (160M) & 0.00 & 0.00 & 0.00 & 0.00 & 0.00 & -- & \textbf{32.61} \\
\midrule
Naive FT & 30.32 & 2.40 & 3.83 & 22.17 & 5.29 & 7.83 & 24.78 \\
% Output Layer Freezing (OLF) & 26.94 & 0.66 & 3.61 & 17.77 & 2.13 & -5.01 & 27.61 \\
%IA3 & 0.00 & 0.00 & 0.00 & 0.00 & 0.00 & -2.13 & 30.48 \\
LoRA & 28.97 & 1.02 & 1.74 & 17.97 & 2.42 & \textbf{1.69} & 30.92 \\
mSGM & 28.39 & 1.37 & 2.71 & 17.48 & 3.36 & 2.68 & 29.93 \\
Soft Targets & \textbf{32.67} & \textbf{6.92} & \textbf{6.10} & \textbf{25.39} & \textbf{10.57} & 2.83 & 29.78 \\
% Soft Targets + OLF & 21.13 & 2.45 & 2.57 & 15.43 & 4.40 & -0.73 & 31.88 \\
% mSGM + OLF & 0.97 & 1.18 & 0.21 & 0.78 & 0.51 & -1.09 & 31.52 \\
% \midrule
\bottomrule
    \end{tabular}
  }
\caption{Results for alternatives to naive fine-tuning for transforming Pythia 160M into an MLLM.}
\label{tab:mitigation-results-pythia160m}
\end{table*}
\begin{figure*}[h!]
    \centering
    \begin{subfigure}[b]{0.49\textwidth}
        \centering
        \includegraphics[width=\textwidth]{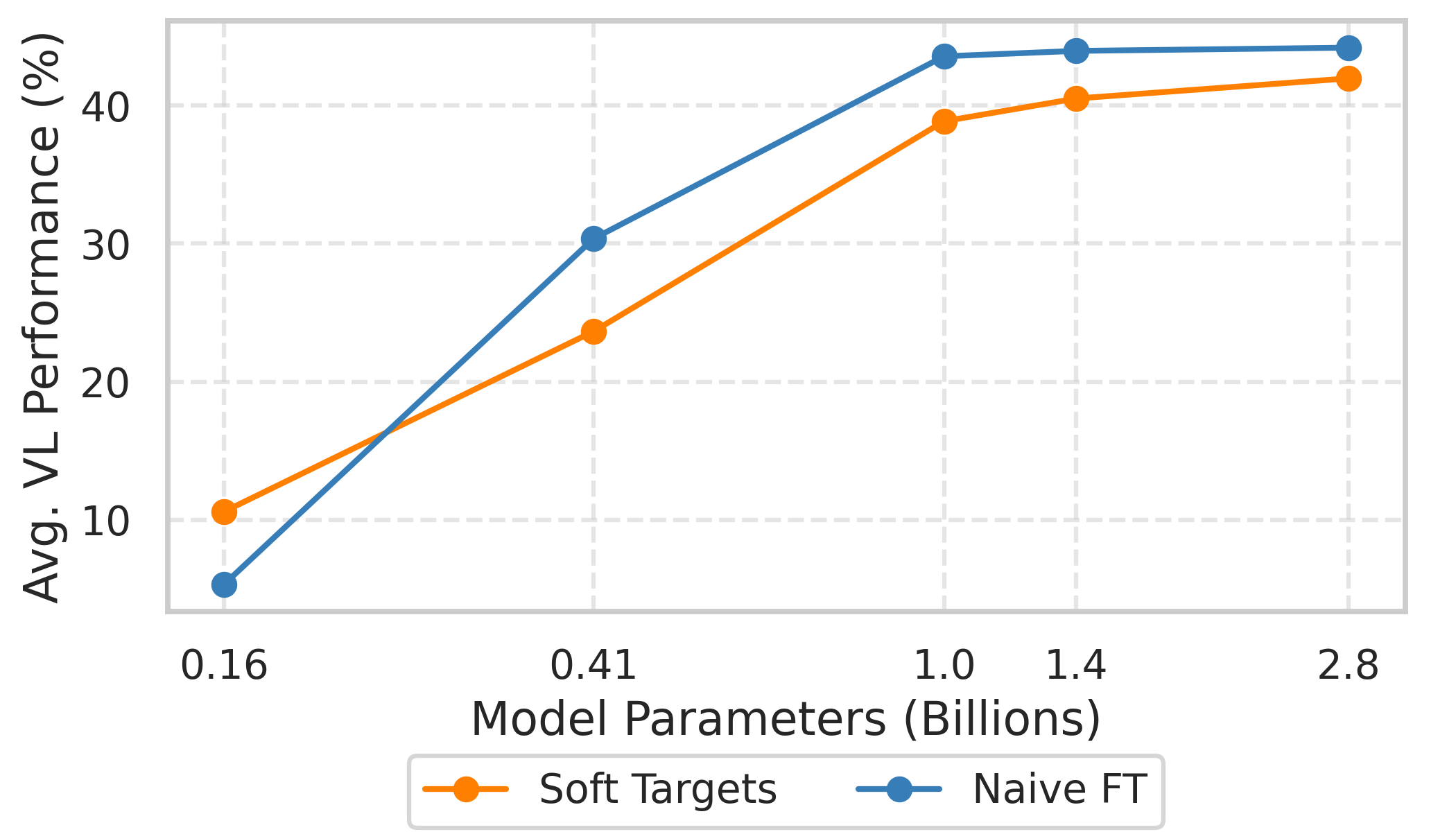}
        \caption{Avg. VL performance with varied model size}
        \label{fig:vl-scaling}
    \end{subfigure}
    \begin{subfigure}[b]{0.49\textwidth}
        \centering
        \includegraphics[width=\textwidth]{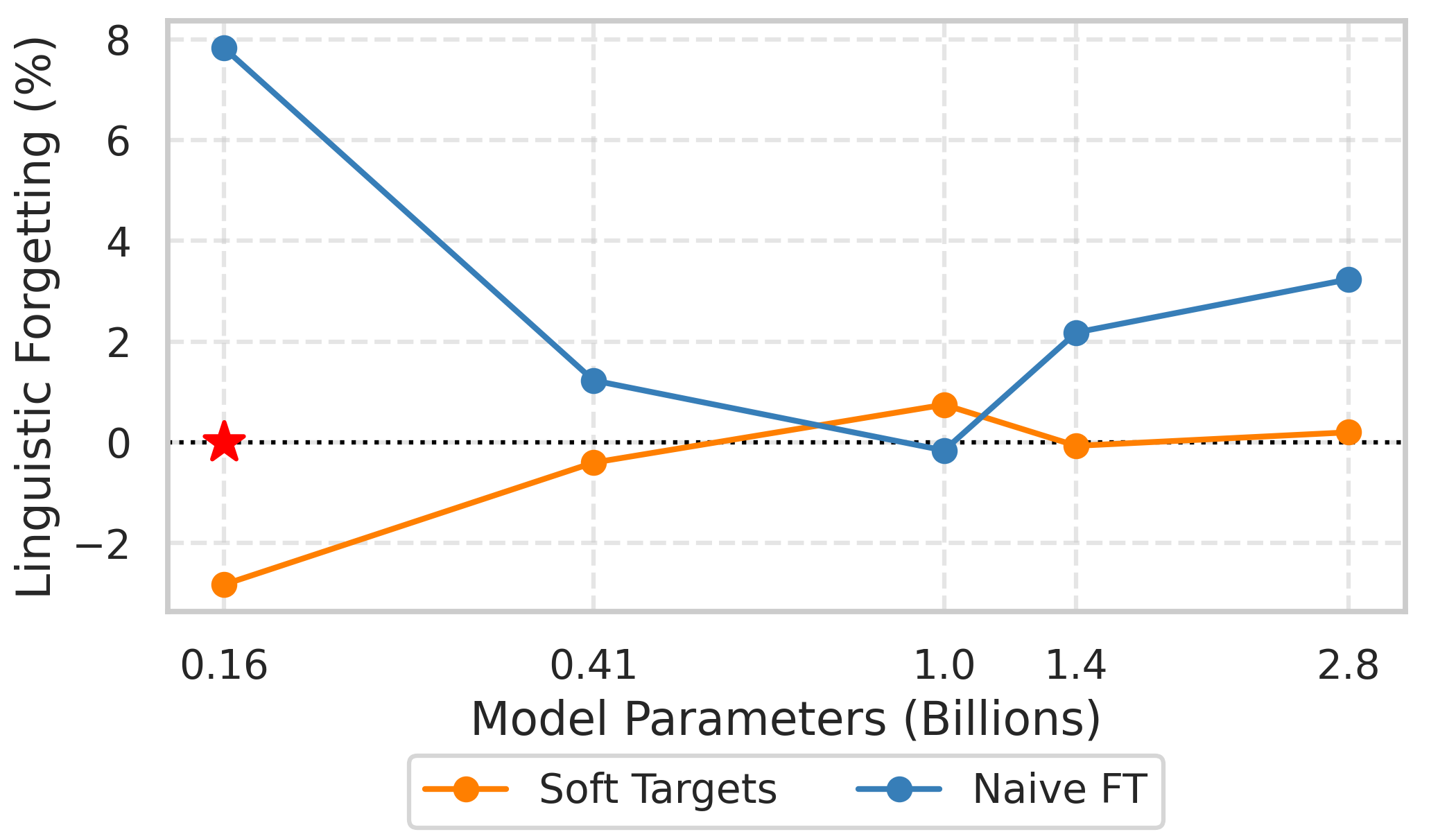}
        \caption{Linguistic forgetting with varied model size}
        \label{fig:nlp-scaling}
    \end{subfigure}
    \caption{\textbf{Vision-language and linguistic forgetting for varying model sizes}. We evaluate the average VL performance and linguistic forgetting for Pythia models in the 160M to 2.8B scale, after training on LLaVA. 
    {In Figure~\ref{fig:nlp-scaling}, \ding{72} refers to the 0 mark where no linguistic forgetting occurs.} 
    Negative forgetting refers to a positive backward transfer in NL performance after multimodal training. 
    }
    \label{fig:model-scaling}
\end{figure*}

\paragraph{Analyzing the Role of Parameter Count.\label{sec:model-scaling-para}}
In \cite{driess2023palm}, larger models had less catastrophic forgetting than smaller ones for NLG/NLU, where their 12B parameter model lost $87.3\%$. In comparison, their 562B parameter model only lost $3.9\%$. The opposite result was found in \cite{luo2023empirical}, where smaller models exhibited \emph{less} forgetting than larger ones. We analyze this phenomenon in the Pythia family of models, where we evaluate naive fine-tuning and Soft Targets, which was the best method for Pythia (160M). Results are given in Fig.~\ref{fig:model-scaling}. Across scales, Soft Targets has zero or negative linguistic forgetting over the naive fine-tuning method used in the original LLaVA paper, with competitive VL performance compared to naive fine-tuning. Soft Targets achieved positive backward transfer in the low model size regime (0.16-0.41B) and no forgetting for larger sizes ($>0.41$B). In contrast, naive fine-tuning had severe forgetting in the 0.16-0.41B regime and reduced forgetting when the model size exceeded 1B parameters.

%%%%%%%%%%%%%%%%%%%%%%%%%%
\subsection{Continually Learning VL Tasks}
\vspace{-0.5em}
We next turn to continual learning on each of the VL datasets used to train LLaVA 1.5, where we group the datasets based on the task type. The task order is provided in Table~\ref{tab:cl-setup}, with additional details in Appendix~\ref{sec:dataset_prep}. Note that there is no evaluation dataset associated with task 2, but we still measure performance on task 1. Given our limited computational budget, we exhaustively studied our CL methods only for the smaller-scale Pythia 410M LLM, and then we evaluated the best-performing mitigation method for all of the Pythia LLMs (from 160M to 1.4B parameters). {Our goal in this setting is continually learn new VL tasks, while mitigating both linguistic forgetting as well as forgetting previously learned VL tasks.}

\begin{table*}[t]
%\vspace{-1em}
  \centering
  \resizebox{\linewidth}{!}{
    \begin{tabular}{l|rr|rr|rr|rr}
     \toprule
     \textbf{Model} & \multicolumn{2}{c|}{\textbf{Task 2 (Instruct)}} & \multicolumn{2}{c|}{\textbf{Task 3 (VQA)}} & \multicolumn{2}{c|}{\textbf{Task 4 (OCR)}} & \multicolumn{2}{c}{\textbf{Task 5 (Ref)}} \\
     & \textbf{VL (A $\uparrow$)} & \textbf{NL ($\Delta \downarrow$)} & \textbf{VL (A $\uparrow$)} & \textbf{NL ($\Delta \downarrow$)} & \textbf{VL (A $\uparrow$)} & \textbf{NL ($\Delta \downarrow$)} & \textbf{VL (A $\uparrow$)} & \textbf{NL ($\Delta \downarrow$)} \\
     \midrule
Naive-FT & - & \textbf{0.58} & \textbf{44.22} & 12.21 & \textbf{16.67} & 4.95 & 0.48 & 7.70 \\
\midrule
Soft Targets & - & 0.81 & 0.16 & 14.67 & 10.23 & 5.41 & 0.31 & 10.90 \\
LoRA & - & 1.38 & 37.46 & 1.23 & 14.03 & 2.50 & 9.59 & 4.36 \\
mSGM & - & 1.11 & 36.31 & 1.57 & 11.69 & \textbf{1.40} & 0.32 & 6.32 \\
Rehearsal \((1\%)\) & - & \textbf{0.58} & 37.74 & 10.90 & 3.47 & 7.41 & 3.55 & 7.65 \\
mSGM + Reh. \((1\%)\) & - & 1.11 & 35.28 & \textbf{0.73} & 12.38 & 2.25 & \textbf{10.21} & \textbf{2.77} \\

     \bottomrule
    \end{tabular}
  }
  \caption{Continually learning LLaVA Tasks with Pythia 410M. We report task-wise {averaged} accuracy and forgetting of each mitigation method across VL and NLU/NLG tasks, where we evaluate all test sets associated with the tasks seen up to the current task.
  }
  \label{tab:vl_nl_acc}
\end{table*}

\begin{figure*}[t]
    \centering
        \begin{subfigure}[b]{0.49\textwidth}
        \centering
        \includegraphics[width=\textwidth]{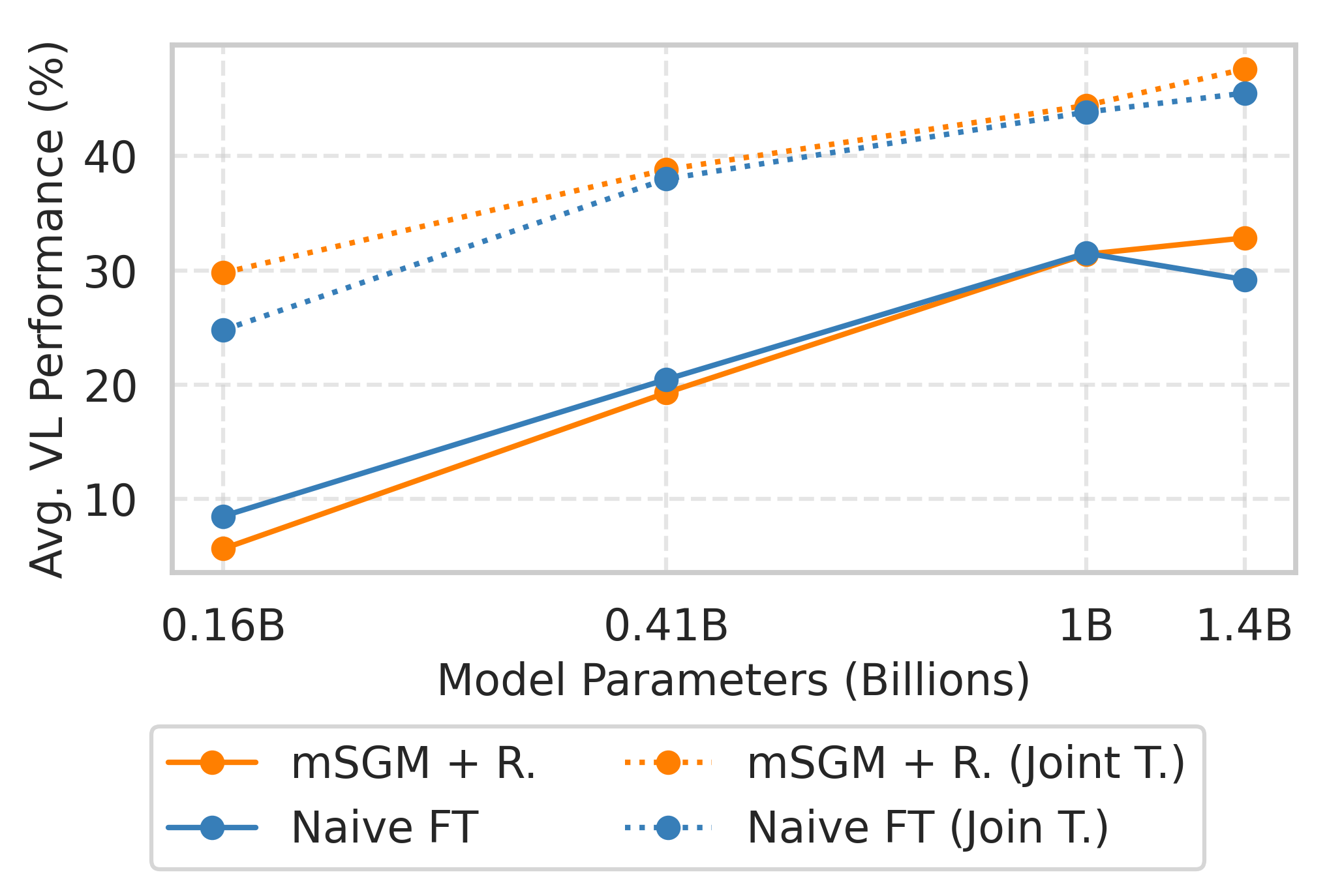}
        \caption{Task averaged VL Performance with varied model size}
        \label{fig:continual-vl-scaling}
    \end{subfigure}
    \begin{subfigure}[b]{0.49\textwidth}
        \centering
        \includegraphics[width=\textwidth]{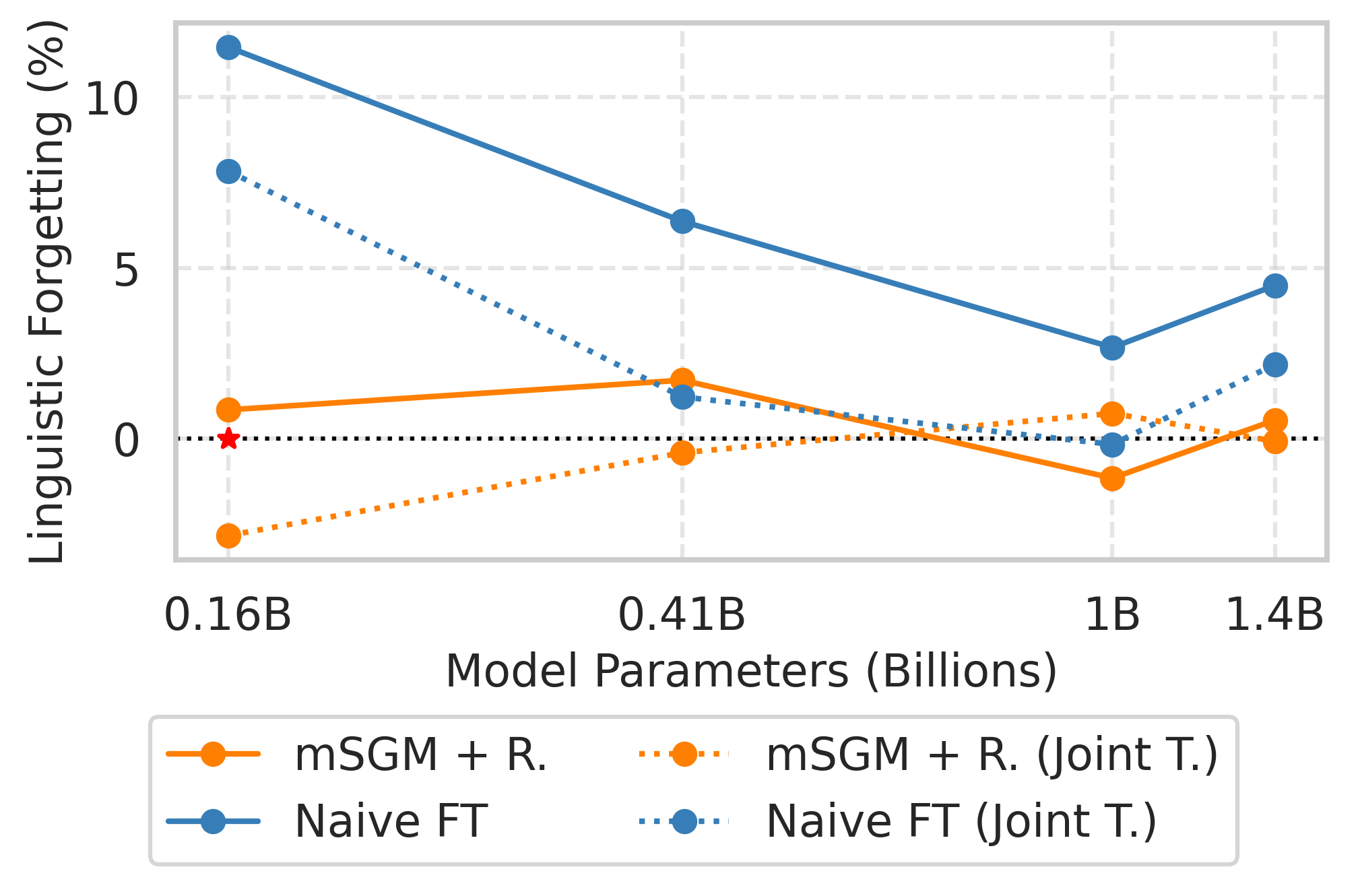}
        \caption{Task averaged Linguistic Forgetting with varied model size}
        \label{fig:continual-nlp-scaling}
    \end{subfigure}
    \caption{\textbf{Continual Learning LLaVA Tasks:} Task averaged Vision-language performance and Linguistic forgetting for varying model sizes. We evaluate mSGM + Rehearsal \((1\%)\) and LLaVA Naive-FT (solid lines for both) on the Continual Learning setup, with varying base LLMs: Pythia models from 160M to 1.4B scale. For each scale, we report the average VL performance or linguistic forgetting across all tasks (task averaged). Performance of joint training  (Joint T.) is shown as an upper-bound (dotted lines) for both Naive FT and mSGM + Rehearsal.
    }
    \label{fig:continual-model-scaling}
    %\vspace{-1.5em}
\end{figure*}

\paragraph{Results for Pythia 410M.}
\begin{wraptable}{r}{0.52\textwidth}
    \vspace{-2.125em}
    \centering
    \resizebox{0.9\linewidth}{!}{%
      \begin{tabular}{cccr}
      \toprule
      \textbf{Task Type} & \textbf{Task} & \textbf{Data} & \textbf{Size} \\
      \midrule
      Pre-Training    & 1 & LLM Pre-Training* & -- \\
      \midrule
      Instruct Tuning & 2  & CC-LAION-SBU & 558K \\
                      &    & LLaVA-Inst, ShareGPT & 198K \\
      \midrule
      VQA & 3 & VQA2 & 83K \\
      (OE \& OK) &   & OKVQA & 9K \\
                 &   & A-OKVQA & 66K \\
                 &   & GQA & 72K \\
      \midrule
      VQA (OCR) & 4 & OCRVQA & 80K \\
                &   & TextCaps & 22K \\
      \midrule
      Referential & 5 & RefCOCO & 48K \\
      Grounding   &   & VisualGenome & 86K \\
      \bottomrule
      \end{tabular}
    }
    \caption{\textbf{Continual LLaVA Setup.} The LLaVA~1.5 data mixture is split into groups of vision-language (VL) tasks. OE \& OK refers to open-ended and outside-knowledge VQA.}
    \label{tab:cl-setup}
    \vspace{-3em}
\end{wraptable}
Results for Pythia 410M are given in Table~\ref{tab:vl_nl_acc}. In terms of NLU/NLG forgetting, all mitigation methods showed efficacy in reducing forgetting across the sequence of tasks. Overall, mSGM with and without rehearsal achieves the least linguistic forgetting while maintaining the highest VL performance across all baselines. 
Regarding VL performance, naive fine-tuning achieves better performance at the cost of high linguistic forgetting. In contrast, mSGM achieves competitive VL performance and sometimes even surpasses it, e.g., for RefCOCO, an especially challenging VL task. {We also report raw accuracies across all tasks in Appendix~\ref{sec:raw_task_accuracies_cl}.}

\paragraph{Model scaling results.} Model scaling results are given in Fig.~\ref{fig:continual-model-scaling}. We compare mSGM with Rehearsal, the best method identified for Pythia 410M with naive fine-tuning. Naive fine-tuning leads to a sharp and consistent drop in task averaged NLU/NLG performance, across all model scales. In contrast, mSGM with Rehearsal has little to no NLU/NLG forgetting across all tasks and multiple model scales. For Pythia 1B, mSGM with Rehearsal achieves positive backward transfer for NLU/NLG datasets as the VL tasks are learned, which is rarely observed in CL. For the 160M parameter model, naive fine-tuning suffers large amounts of forgetting compared to mSGM. Across all scales, naive fine-tuning exhibits greater losses of NLU/NLG performance, unlike mSGM. In general, VL performance for mSGM rivals naive fine-tuning across scales. And in the case of the largest 1.4B scale, exceeds the task averaged performance of naive fine-tuning.

%%%%%%%%%%%%%%%%%%%%%%%%%%%%%%%%%%%%
\section{Discussion}
This work presents one of the first studies of linguistic forgetting in MLLMs, particularly for open-source models with modest parameter counts ($<7$B). We show that the degree of linguistic forgetting typically reduces with model scale, and is far more severe for NLG tasks compared to NLU. Additionally, we demonstrate that instruction-tuned base LLMs are more robust to linguistic forgetting from multimodal training than non instruction-tuned base LLMs. We pioneer treating MLLM creation as a CL problem, and show that CL methods effectively mitigate linguistic forgetting while minimally hindering VL accuracy. In our experiments, our best mitigation method far outperforms the naive LLaVA recipe in terms of linguistic forgetting, while maintaining competitive VL performance. We show that this benefit holds across model scales. We pioneer CL for MLLMs, and establish strong baselines for this task. In fact, we observe that besides maintaining competitive VL performance with naive LLaVA training, our mitigation methods achieve near zero and in some cases even negative linguistic forgetting. This suggest that our mitigation methods achieve a positive backward transfer of linguistic ability after multimodal training. Given the essential nature of multimodal abilities for many applications, our research highlights that naive multimodal fine-tuning can significantly degrade prior linguistic abilities. Our findings underscore the importance of developing more robust and capable mitigation approaches, showcasing the applicability of CL techniques to the fine-tuning of foundation models. We aim to inspire future research in this direction, ultimately contributing to the advancement of more resilient and versatile MLLMs.

%%%%%%%%%%%%%%%%%%%%%
\section{Future Work}

In this work, we rigorously studied catastrophic forgetting in MLLMs trained with the LLaVA recipe. While our findings provide valuable insights, several avenues remain open for future exploration. 

We studied LLaVA, but we assume our findings would apply to similar models~\citep{liu2024visual,liu2023improved,lin2023vila}. Of the PET methods, we only studied LoRA; however, other PET techniques may be more effective~\citep{han2024parameter}. Future work could study other methods and investigate their ability to reduce forgetting in MLLMs. 

We experimented with 9 MLLMs and our experiments were computationally demanding. Therefore, we could not study all of the methods for every LLM, meaning that using the best approach for only a small LLM may not have ultimately been the best for every LLM. Likewise, our limited computational resources meant we could not study variability in performance for all permutations of the tasks in our CL experiments. Future work might investigate how learning more tasks affects forgetting in MLLMs.

We only studied models that integrate vision and language, but some MLLMs include additional modalities~\citep{driess2023palm}. It is not clear if our methods will be as effective for those. It would be interesting to study MLLMs that also have multimodal outputs for multimodal dialogue in CL settings. FROMAGe studied this in a non-CL setting with frozen LLM~\citep{koh2023grounding}. Exploration of different CL methods is an exciting future direction to build a multimodal chatbot or agent that can continually and seamlessly interact with users.

Following the LLaVA training recipe, we only trained models with a single pass through the VL datasets. We hypothesize that the larger MLLMs may be under-trained based on our results, considering that their VL performance plateaued compared to smaller models. Adopting this standard does help to ensure a consistent and fair analysis; however, for practical purposes of creating better MLLMs, it would be ideal to maximize VL performance while minimizing NLU/NLG forgetting.

Due to our limited compute budget, we could not study MLLMs with more than 7B parameters. It would be interesting to examine how model size impacts catastrophic forgetting in larger LLMs. For instance, PaLM-E (12B) exhibits $87.3$\% forgetting whereas the largest model, PaLM-E (562B), exhibits only $3.9$\% forgetting. Scaling our methods to larger models is an exciting direction for future work. 

%%%%%%%%%%%%%%%%%%%%%%%
%Acknowlegement

\ifthenelse{\boolean{addAck}}{
\section*{Acknowledgments}
This work was supported in part by NSF awards \#2326491 and \#2317706. The views and conclusions contained herein are those of the authors and should not be interpreted as representing any sponsor's official policies or endorsements.
}

\bibliography{custom}
\bibliographystyle{collas2025_conference}

\clearpage
\appendix

\begin{strip}
\begin{center}
    {\Large{\textbf{Appendix}}}
\end{center}
\end{strip}

\begin{center}
    {\Large{Appendix}}
\end{center}

We organize implementation details and additional supporting experimental findings as follows: 
\begin{itemize}
    \item Appendix~\ref{sec:reproducibility} includes reproducibility and code release information.
    \item Appendix~\ref{sec:hardware-and-compute} provides details on compute resources (GPU) and training time.
    \item Appendix~\ref{sec:dataset_prep} provides details on the datasets used in this paper.
    \item Appendix~\ref{sec:supporting_exp} summarizes findings of additional supporting experiments and ablation studies.
\end{itemize}

%%%%%%%%
\section{Reproducibility and Code Release}
\label{sec:reproducibility}
All datasets and pre-trained models utilized in this work are widely recognized and publicly available. To facilitate replication of our experiments, we will create a GitHub repository with codes and a dedicated website. 
The code will be released under an MIT license. We will also release our trained MLLMs (weights and configurations).
These resources will provide the exact sequence of examples used for each continual learning experiment. Additionally, the Appendix contains a comprehensive description of the hyperparameters and optimizer settings employed in each experiment.

%%%%%%%%
\section{Hardware and Computational Budget}
\label{sec:hardware-and-compute}

We trained a total of 9 MLLMs on 6 datasets, over 170 experiments on 4 A100 40GB GPU, with an aggregated compute time of \(\sim3400\) (wall clock) hours (\(\sim141\) days).

\subsection{Hyperparameter Search}

\subsubsection{LLaVA1.5 Setting}
To reproduce and directly compare against the LLaVA 1.5 protocol, we keep the explicit training configurations the same (as mentioned in Appendix~\ref{app:exp_settings}).

\subsubsection{LoRA\label{sec:lora-hparam}}
We train several different LoRA settings on the smallest Pythia-160M LLM for tractability and compare the resulting VL and NLU/NLG performances. We vary the a) target modules (between 1) all linear layers, and 2) key, query, and value projection layers only), b) LoRA ranks in the range of \(1/4 - 1/2\) of the full rank of the model, rank stabilized LoRA~\citep{kalajdzievski2023rank}, and larger alpha values (\(16\) instead of \(8\) as default). Table~\ref{tab:lora_variants} shows the comparisons by training on the LLaVA setting. 

\begin{table*}[h!]
  \caption{\textbf{Analysis of LoRA Ranks and Configuration:} We train the Pythia 160M model with a varying set of ranks and configurations.}
  \label{tab:lora_variants}
  \centering
  \resizebox{\linewidth}{!}{
    \begin{tabular}{l|rrrr|r|rr}
     \toprule
     \textbf{Model} & \multicolumn{4}{c|}{\textbf{Vision-Language (VL)}} & \textbf{VL Avg.} & \multicolumn{2}{c}{\textbf{NL Avg.}} \\
     & \textbf{VQAv2} & \textbf{TextVQA OCR} & \textbf{TextVQA Pure} & \textbf{GQA} & Acc $\uparrow$ & $\Delta \downarrow$ & Acc $\uparrow$ \\
     \midrule
Original LLaVA & 30.32 & 2.40 & 3.83 & 22.17 & 5.29 & 7.83 & 24.78 \\
Language Only LLM & 0.00 & 0.00 & 0.00 & 0.00 & 0.00 & - & 32.61 \\
\midrule
LoRA (1/2 Full Rank, Higher Alpha) & 28.72 & 1.05 & 2.67 & 19.73 & 2.84 & 9.33 & 23.28 \\
LoRA (1/2 Full Rank, RSLoRA) & 28.97 & 1.02 & 1.74 & 17.97 & 2.42 & 1.69 & 30.92 \\
LoRA (1/4 Full Rank) & 24.64 & 0.93 & 1.41 & 15.04 & 2.11 & 11.64 & 20.97 \\
LoRA (1/4 Full Rank, Higher Alpha) & 6.46 & 0.68 & 0.55 & 2.44 & 1.04 & 2.53 & 30.08 \\
LoRA (1/2 Full Rank) & 0.13 & 0.20 & 0.10 & 0.00 & 0.00 & - & - \\
LoRA (1/2 Full Rank, RSLoRA, KQV Target) & 0.00 & 0.00 & 0.00 & 0.00 & 0.00 & - & - \\
     \bottomrule
    \end{tabular}
  }
\end{table*}

\subsubsection{Soft Targets\label{sec:soft-target-hparam}}
We train the LLaVA recipe with the soft targets with varying alpha \(\alpha \in \{0.001,0.01, 0.1\}\), and report results in Figure ~\ref{tab:soft-targets-hparam}. The Pythia 160M model is used for this tuning.

\begin{table*}[h]
  \caption{\textbf{Selecting \(\alpha\) for Soft Targets}. We train the Pythia 160M model with Soft Targets by varying the \(alpha \in \{0.001,0.01, 0.1\}\).}
  \label{tab:soft-targets-hparam}
  \centering
  \resizebox{\linewidth}{!}{
    \begin{tabular}{l|rrrr|r|rr}
     \toprule
     \textbf{Model} & \multicolumn{4}{c|}{\textbf{Vision-Language (VL)}} & \textbf{VL Avg.} & \multicolumn{2}{c}{\textbf{NL Avg.}} \\
     & \textbf{VQAv2} & \textbf{TextVQA OCR} & \textbf{TextVQA Pure} & \textbf{GQA} & Acc $\uparrow$ & $\Delta \downarrow$ & Acc $\uparrow$ \\
     \midrule
Language Only LLM & 0.00 & 0.00 & 0.00 & 0.00 & 0.00 & - & 32.61 \\
\midrule
Soft Targets \((\alpha=0.1)\) & 3.38 & 1.19 & 1.32 & 1.76 & 1.62 & 0.67 & 31.95 \\
Soft Targets \((\alpha=0.01)\) & 32.67 & 6.92 & 6.10 & 25.39 & 10.57 & 2.83 & 29.78 \\
Soft Targets \((\alpha=0.001)\) & 25.12 & 2.17 & 1.73 & 14.84 & 3.49 & 4.97 & 27.64 \\
\midrule
Original LLaVA & 30.32 & 2.40 & 3.83 & 22.17 & 5.29 & 7.83 & 24.78 \\
\bottomrule
    \end{tabular}
  }
\end{table*}

\section{Dataset Preparation}
\label{sec:dataset_prep}
\subsection{Continual LLaVA}
\textbf{All datasets are primarily in the English language.} However, some small samples of other languages such as Chinese can be found in the \textit{ShareGPT} subset of LLaVA Instruct~\citep{liu2023llava}, for instance.
The preprocessing steps required for splitting the LLaVA 1.5 Instruct json into component datasets, to build our continual learning setting, are detailed below:
The primary LLaVA instruct json \texttt{llava\_v1\_5\_mix665k.json}~\citep{liu2023llava} is split into component datasets based on the data entries as follows:
\begin{itemize}
    \item \textbf{LLaVA Instruct~\citep{liu2023llava}}: The first 157,712 entries are directly assigned to the LLaVA subset, per manual verification.
    \item \textbf{ShareGPT (SG)~\citep{liu2023improved}}: Entries without an associated image are categorized under ShareGPT.
    \item \textbf{GQA~\citep{hudson2019gqa}, OCR-VQA, TextVQA~\citep{singh2019towards}, Visual Genome (VG)~\citep{krishna2017visual}}: Entries associated with images from the GQA, OCR-VQA, TextVQA, and \texttt{VG\_100K} are mapped correspondingly to their respective datasets.
    \item \textbf{COCO-Based Datasets}: For images related to the COCO dataset, further categorization is performed based on the prompts:
        \begin{itemize}
            \item \textbf{OKVQA}~\citep{marino2019ok}: If the prompt contains the phrase ``Answer the question using a single word or phrase'' and the image ID matches those in the OKVQA set.
            \item \textbf{VQA-V2}~\citep{goyal2017making}: If the above condition is not met, the entry is categorized under VQA-V2.
            \item \textbf{A-OKVQA}~\citep{schwenk2022okvqa}: If the prompt contains multiple choice options (`A.', `B.', `C.').
            \item \textbf{RefCOCO}~\citep{kazemzadeh2014referitgame}: If the prompt requests bounding box coordinates or a short description of a region.
        \end{itemize}
\end{itemize}

The splitting process categorizes the dataset into distinct subsets as shown in Table~\ref{tab:dataset_split}, which are saved as separate JSON files in the specified directory. The number of entries in each subset is printed for verification. 

\begin{table}[h!]
  \caption{Dataset Split generated based on LLaVA 1.5 Instruct composition~\cite{liu2023improved}.}
  \centering
  %\resizebox{\linewidth}{!}{
     \begin{tabular}{lr}
     \hline
     \textbf{Subset} & \textbf{Sample Size} \\
     \hline
     LLaVA & 157,712 \\
     SG & 40,688 \\
     GQA & 72,140 \\
     OKVQA & 8,994 \\
     OCRVQA & 80,000 \\
     A-OKVQA & 66,160 \\
     TextCaps & 21,953 \\
     RefCOCO & 48,447 \\
     VG & 86,417 \\
     VQA-V2 & 82,787 \\
     \hline
    \end{tabular} %}
  \label{tab:dataset_split}
\end{table}

After splitting, the datasets are then combined, based on the grouping in Table \ref{tab:cl-setup}, to form the respective sequence of Tasks 1 to 5. 

\subsection{Evaluation Dataset Splits}
We use the \emph{slim}\footnote{\url{https://github.com/TRI-ML/vlm-evaluation}} versions of all VQAv2, GQA, Text-VQA and RefCOCO datasets for evaluation, as provided within Prismatic-VLMs~\citep{karamcheti2024prismatic}. All \emph{slim} versions of the evaluation sets contain \(1024\) examples each, and we use the provided index splits for testing.

\section{Additional Experiments}
\label{sec:supporting_exp}

\subsection{Analysis of Forgetting across NLU and NLG tasks\label{sec:analysis-nlu-nlg-forgetting}}
\begin{figure*}[!h]
    \centering
    \includegraphics[width=0.775\linewidth]{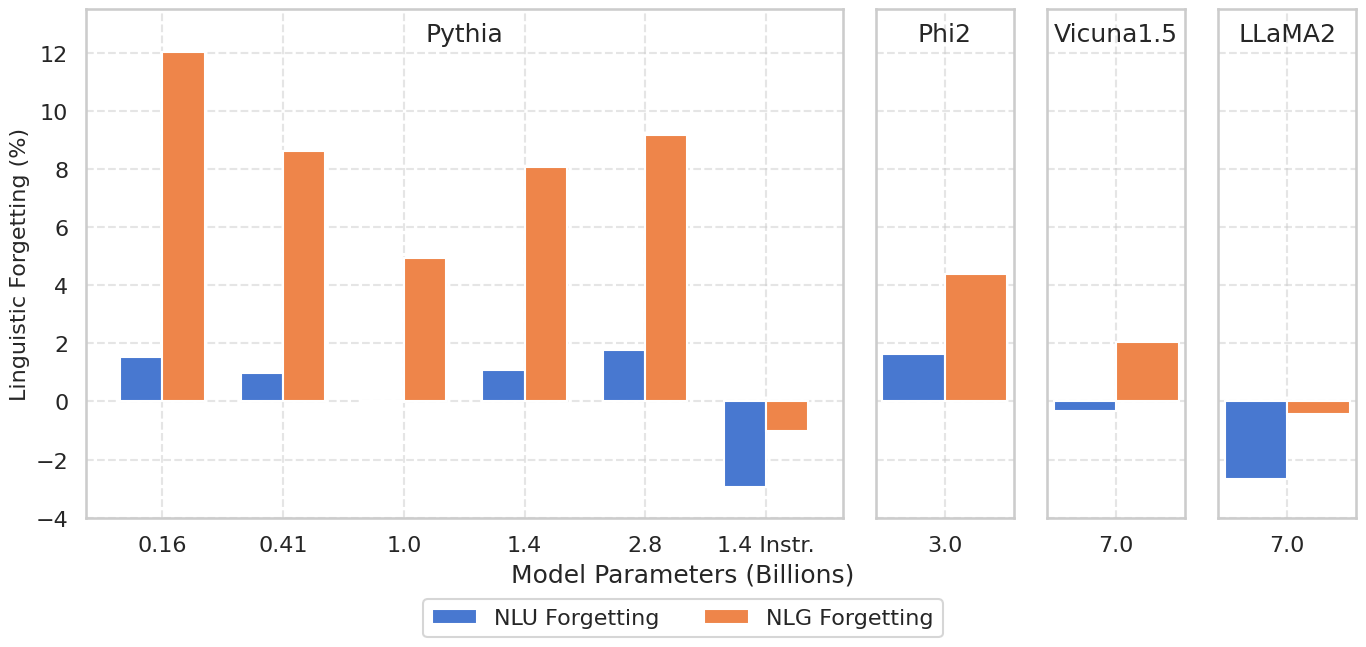}
    \caption{\textbf{Linguistic forgetting by NLU and NLG tasks}: For the LLaVA setting, we look at different model scales and families, and show the degree of linguistic forgetting by both NLU and NLG tasks separately.}
    \label{fig:forgetting-nlu-nlg}
\end{figure*}

To understand the composition of linguistic forgetting on the LLaVA setting, we look at the NLU and NLG forgetting for 9 different LLMs with varying scales: Pythia family 160M to 2.8B, Phi2 3B, Vicuna1.5 7B and LLaMA2 7B. Figure ~\ref{fig:forgetting-nlu-nlg} shows these results. Here, the NLU and NLG averages are computed as the simple mean. We can observe a clear trend of higher forgetting for the NLG dataset (Lambada), compared to forgetting for NLU. Another trend we note is that forgetting typically reduces with higher model scales, for both NLG and NLU. 
% \subsection{Analyzing Components of mSGM}

\begin{figure*}[h]
    \centering
    \includegraphics[width=0.70\linewidth]{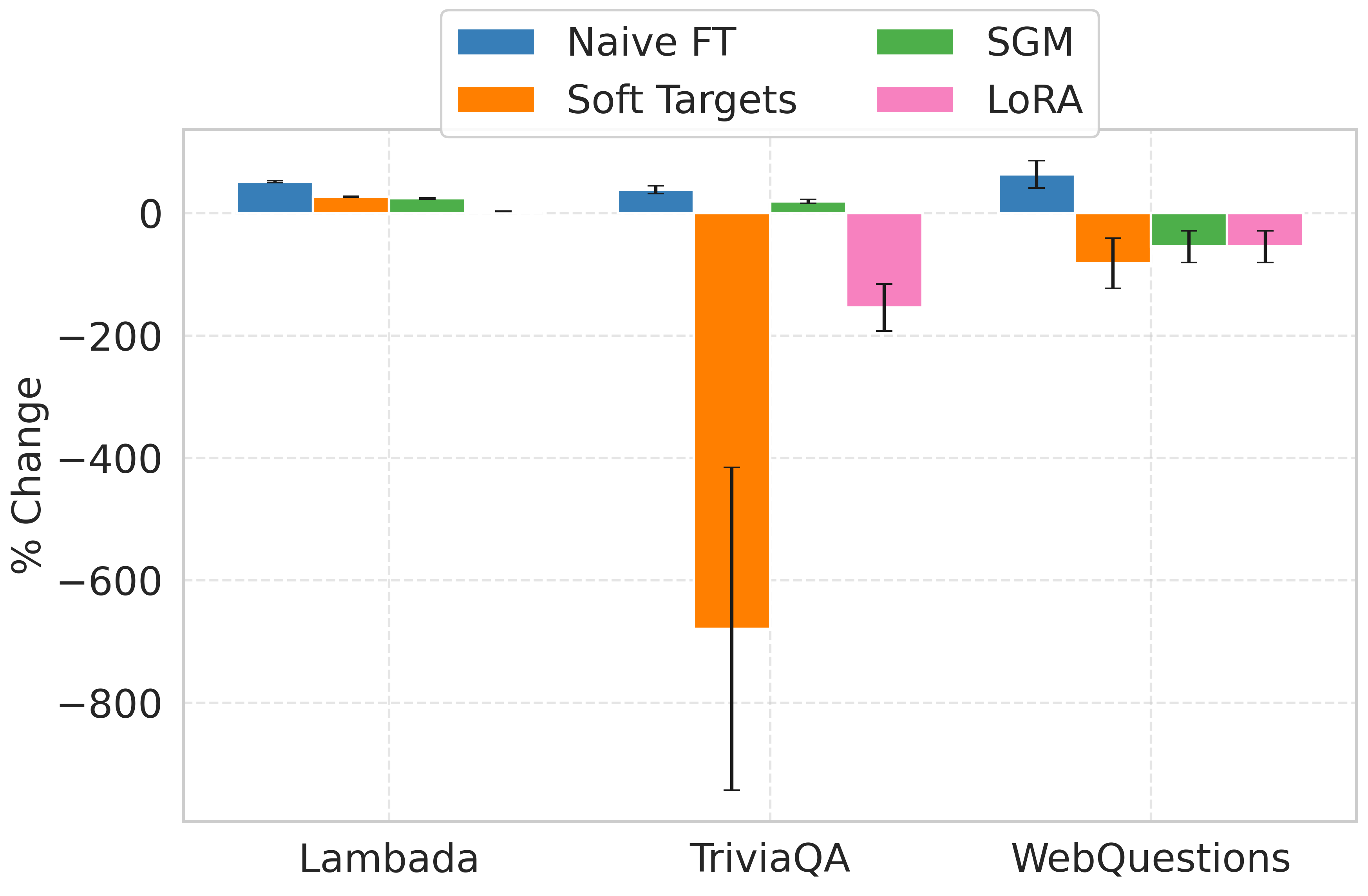}
    \caption{{\textbf{Linguistic forgetting across various NLG tasks}: For the LLaVA setting, we look at the degree of linguistic forgetting across multiple NLG datasets: WebQuestions~\citep{berant2013semantic}, TriviaQA~\citep{joshi2017triviaqa} and Lambada~\citep{paperno2016lambada}. In order to use a single metric across WebQuestions and TriviaQA (exact-match (EM)) and Lambada (accuracy), we use the ``\% Change" metric \(\Delta' = \frac{w_{\mathrm{(Base\; LLM)}} - w_{(\mathrm{MLLM\;  Method})}}{w_{\mathrm{(Base\; LLM)}}} \times 100\%\). The NLG forgetting across all methods is reported at the 160M scale. The error bars show the degree of standard error in the NLG evaluations. Lower values are better. Negative \(\Delta'\) values indicate positive backward transfer. }}
    \label{fig:forgetting-nlg-extra}
\end{figure*}

{\subsection{Analysis of Linguistic Forgetting across multiple NLG tasks}\label{sec:multiple_nlg_analysis}}
{In Figure ~\ref{fig:forgetting-nlg-extra}, we analyze the degree of linguistic forgetting across multiple NLG datasets: WebQuestions~\citep{berant2013semantic}, TriviaQA~\citep{joshi2017triviaqa} and Lambada~\citep{paperno2016lambada}. We show for various methods using the LLaVA recipe at the 160M scale, that the degree of NLG forgetting is significant (between 40\% and to 64\%) for the Original LLaVA (Naive FT) across various NLG datasets, and is not specific to Lambada~\citep{paperno2016lambada}. Additionally, we observe that our best method - Soft Targets outperforms other mitigation approaches by large margins in almost all cases, towards mitigating the degree of NLG forgetting. While LoRA and mSGM, appear to help mitigate forgetting, we hypothesize that the retention of lower probability logits in the soft targets approach is critical for retaining linguistic generation abilities.}

\subsection{Analyzing Task Order Sensitivity}
We report the performance of the mSGM + Reh. and the Original LLaVA (Naive FT) for varying task orders in Tables ~\ref{tab:task_order_sensitivity_1} and \ref{tab:task_order_sensitivity_2}. Naive FT, unlike mSGM + Reh., retains little to no VL performance by Task 5, with large NL forgetting.

\begin{table*}[!h]
  \caption{\textbf{Task order 1:} Task-wise Accuracies and Forgetting of Each Mitigation Method across VL and NL tasks}
  \label{tab:task_order_sensitivity_1}
  \centering
  \resizebox{\linewidth}{!}{
    \begin{tabular}{l|cc|cc|cc|cc}
     \toprule
     \textbf{Model} & \multicolumn{2}{c|}{\textbf{Task 2 (Instruct)}} & \multicolumn{2}{c|}{\textbf{Task 3 (OCR)}} & \multicolumn{2}{c|}{\textbf{Task 4 (Ref)}} & \multicolumn{2}{c|}{\textbf{Task 5 (VQA)}} \\
     & \textbf{VL (A $\uparrow$)} & \textbf{NL ($\Delta \downarrow$)}     & \textbf{VL (A $\uparrow$)} & \textbf{NL ($\Delta \downarrow$)}     & \textbf{VL (A $\uparrow$)} & \textbf{NL ($\Delta \downarrow$)}     & \textbf{VL (A $\uparrow$)} & \textbf{NL ($\Delta \downarrow$)} \\
     \midrule
LLaVA Naive FT & 0.10 & 3.72 & 0.10 & 17.17 & 0.12 & 12.71 & 0.14 & 26.61 \\
\midrule
mSGM + Reh. \((1\%)\) & 0.10 & -0.46 & 0.06 & 1.79 & 0.13 & 1.94 & 0.46 & 4.66 \\

     \bottomrule
    \end{tabular}
  }
\end{table*}

\begin{table*}[!h]
  \caption{\textbf{Task order 2:} Task-wise Accuracies and Forgetting of Each Mitigation Method across VL and NL tasks}
  \label{tab:task_order_sensitivity_2}
  \centering
  \resizebox{\linewidth}{!}{
    \begin{tabular}{l|cc|cc|cc|cc}
     \toprule
     \textbf{Model} & \multicolumn{2}{c|}{\textbf{Task 2 (VQA)}} & \multicolumn{2}{c|}{\textbf{Task 3 (OCR)}} & \multicolumn{2}{c|}{\textbf{Task 4 (Instruct)}} & \multicolumn{2}{c|}{\textbf{Task 5 (Ref)}} \\
     & \textbf{VL (A $\uparrow$)} & \textbf{NL ($\Delta \downarrow$)}     & \textbf{VL (A $\uparrow$)} & \textbf{NL ($\Delta \downarrow$)}     & \textbf{VL (A $\uparrow$)} & \textbf{NL ($\Delta \downarrow$)}     & \textbf{VL (A $\uparrow$)} & \textbf{NL ($\Delta \downarrow$)} \\
     \midrule
LLaVA Naive FT & 32.41 & 15.74 & 0.56 & 11.20 & 0.10 & 6.49 & 0.15 & 18.24 \\
\midrule
mSGM + Reh. \((1\%)\) & 8.90 & 6.39 & 2.82 & 9.62 & 2.28 & 2.54 & 0.44 & 5.07 \\

     \bottomrule
    \end{tabular}
  }
\end{table*}

{\subsection{Task wise accuracies for Instruction-tuning Models}}
{The individual task accuracy breakups for both NL as well as VL tasks are shown in Tables~\ref{tab:instruct_tuned_nl_accuracies}, ~\ref{tab:instruct_tuned_vl_accuracies}. We observe as before instruction-tuned LLaMA 2 (7B) and Pythia (1.4B) models show minimal or even negative linguistic forgetting, indicating positive backward transfer. 
Conversely, base LLMs display greater variability, with base Pythia experiencing more forgetting than base LLaMA 2, possibly due to differences in their pre-training datasets affecting post-multimodal training linguistic performance. Instruction tuning significantly reduces NLG forgetting in Pythia (1.4B) and induces positive backward transfer from VL to NLG, with a similar trend observed in NLU tasks. This suggests that instruction-tuned LLMs exhibit little to no forgetting, likely due to the distributional similarity between text-only instruction tuning tasks and visual-instruction tuning, helping to mitigate catastrophic forgetting during multimodal training.}

\begin{table*}[h]
  \centering
  \caption{\textbf{Natural Language Task Accuracies: Instruction Tuning Analysis between Base LLMs vs LLaVA MLLMs.} Raw accuracies on NL tasks are shown for our instruction‑tuning study. Instr.\ column is checked, if the underlying LLM is instruction‑tuned. Base LLMs are the underlying LLMs used to create the LLaVA MLLM.}
  \label{tab:instruct_tuned_nl_accuracies}
  \resizebox{0.8\linewidth}{!}{%
    \begin{tabular}{l|c|ccccc}
      \toprule
      \textbf{Model}             & \textbf{Instr.} & \textbf{WSC273} & \textbf{Winogrande} & \textbf{ARC‑E} & \textbf{ARC‑C} & \textbf{Lambada} \\
      \midrule
      \multicolumn{7}{l}{\textit{\textbf{Base LLMs}}}                                                                                          \\
      \midrule
      \multirow{2}{*}{Pythia (1.4B)}   & \ding{55}       & 70.70           & 56.51               & 61.74          & 27.47          & 48.98           \\
                                      & \ding{51}       & 59.34           & 50.43               & 48.23          & 26.79          & 33.79           \\
      \midrule
      \multirow{2}{*}{LLaMA2 (7B)}     & \ding{51}       & 85.35           & 69.53               & 75.63          & 43.17          & 64.35           \\
                                      & \ding{55}       & 80.59           & 69.22               & 76.26          & 43.43          & 68.27           \\
      \midrule
      \multicolumn{7}{l}{\textit{\textbf{LLaVA MLLMs}}}                                                                                         \\
      \midrule
      \multirow{2}{*}{Pythia (1.4B)}   & \ding{55}       & 67.40           & 56.20               & 60.98          & 27.47          & 40.91           \\
                                      & \ding{51}       & 64.84           & 54.06               & 51.94          & 25.68          & 34.80           \\
      \midrule
      \multirow{2}{*}{LLaMA2 (7B)}     & \ding{51}       & 85.35           & 68.35               & 75.97          & 45.39          & 62.31           \\
                                      & \ding{55}       & 87.91           & 70.56               & 77.06          & 44.62          & 68.70           \\
      \bottomrule
    \end{tabular}%
  }
\end{table*}

\begin{table*}[h]
  \centering
  \caption{\textbf{Vision‑Language Task Accuracies: Instruction‑tuning Analysis.} Raw accuracies on VL tasks are shown for our instruction‑tuning study. Instr.\ column is checked, if the underlying LLM is instruction‑tuned.}
  \label{tab:instruct_tuned_vl_accuracies}
  \resizebox{0.8\linewidth}{!}{%
    \begin{tabular}{l|c|cccc}
      \toprule
      \textbf{Model}            & \textbf{Instr.} & \textbf{VQA} & \textbf{TextVQA‑OCR} & \textbf{TextVQA‑Pure} & \textbf{GQA} \\
      \midrule
      \multirow{2}{*}{Pythia (1.4B)} & \ding{55}       & 66.17        & 38.49                 & 35.49                  & 46.09        \\
                                     & \ding{51}       & 66.46        & 39.12                 & 34.35                  & 46.88        \\
      \midrule
      \multirow{1}{*}{LLaMA2 (7B)}   & \ding{51}       & 74.50        & 56.28                 & 45.95                  & 56.25        \\
      \bottomrule
    \end{tabular}%
  }
\end{table*}

%%%%%%%%%%%%%%%%%%%%%%%%%%
%%%%%%%%%%%%%%%%%%%%%%%%%%

{\subsection{Task-wise Accuracies during Continual Training}\label{sec:raw_task_accuracies_cl}}
{In Table~\ref{tab:vl_acc_long}, we report the raw accuracies during continual training. We observe that Original LLaVA (Naive FT) heavily biases on the first task that is learned i.e. VQAv2, and consequently performs strongly on the VQA evaluations (VQAv2 and GQA) as well as the TextVQA-Pure benchmark, which is also similarly a VQA task.  However, it is unable to retain performance on OCR tasks after learning RefCOCO. Additionally, it completely loses its plasticity and fails to learn the RefCOCO task. As before heavily biases on the VQA tasks. The mSGM + Rehearsal(\(1\%\)) approach on the other hand, learns and retains both OCR and VQA performances more effectively than all other approaches, while simultaneously also outperforming all other approaches on the final RefCOCO task - the most complex task in the LLaVA recipe for MLLMs to learn in our tests. This evidences the ability to effectively manage both plasticity and mitigating forgetting during the continual LLaVA training.}

{\begin{longtable}{l|ccccc}
\caption{{\textbf{Raw VL Accuracy during Continual LLaVA}. We report the raw accuracies (\%) on the five vision‐language tasks: VQAv2 (3), GQA (3), TextVQA-OCR (4), TextVQA Pure (4) and RefCOCO (5), evaluated after continual training up to each stage  (2: Instruct, 3: VQA, 4: OCR, 5: Ref). The index of the evaluated vision-language tasks show which stage they are expected to evaluate for. As before, stage 1 refers to the LLM pre-training stage. Rows (“Trained on \(\downarrow\)”) indicates the last task stage seen during training and columns (“Evaluated \(\rightarrow\)”) denote the task under test. For ``Instruct (2)", the evaluation accuracies are reported where available. Each sub-table shows one mitigation method; stable accuracy across rows demonstrates effective knowledge retention, whereas drops on earlier tasks after later‐stage training reveal catastrophic forgetting. In reporting the task-averaged accuracies in our main results, we consider only the average of VL tasks it is trained on. }}\label{tab:vl_acc_long} \\
\toprule
\multirow{2}{*}{\textbf{Trained on} $\downarrow$} & \multicolumn{5}{c}{\textbf{Evaluated} $\rightarrow$} \\
 & \textbf{VQAv2 (3)} & \textbf{GQA (3)} & \textbf{TextVQA-OCR (4)} & \textbf{TextVQA-Pure (4)} & \textbf{RefCOCO (5)} \\
\midrule
\endfirsthead
\caption[]{\textbf{(continued)} Raw VL Accuracy during Continual LLaVA} \\
\toprule
\multirow{2}{*}{\textbf{Trained on} $\downarrow$} & \multicolumn{5}{c}{\textbf{Evaluated} $\rightarrow$} \\
 & \textbf{VQAv2 (3)} & \textbf{GQA (3)} & \textbf{TextVQA-OCR (4)} & \textbf{TextVQA-Pure (4)} & \textbf{RefCOCO (5)} \\
\midrule
\endhead
\bottomrule
\endfoot
\midrule
\multicolumn{6}{l}{\textbf{Original LLaVA}} \\
\midrule
Instruct (2) & --- & --- & --- & --- & --- \\
VQA (3) & 51.44 & 38.77 & 9.84 & 10.78 & 0.00 \\
OCR (4) & 45.13 & 30.47 & 9.86 & 11.96 & 0.00 \\
Ref (5) & 46.84 & 30.57 & 3.90 & 10.57 & 0.00 \\
\midrule
\multicolumn{6}{l}{\textbf{Soft Targets (ST)}} \\
\midrule
Instruct (2) & 0.45 & 0.00 & 1.55 & 1.15 & 0.00 \\
VQA (3) & 0.39 & 0.00 & 0.31 & 0.41 & 0.00 \\
OCR (4) & 35.64 & 26.07 & 8.03 & 5.00 & 0.00 \\
Ref (5) & 4.92 & 2.15 & 0.21 & 1.26 & 0.00 \\
\midrule
\multicolumn{6}{l}{\textbf{mSGM}} \\
\midrule
Instruct (2) & 0.10 & 0.00 & 0.13 & 0.00 & 0.00 \\
VQA (3) & 43.53 & 31.15 & 12.63 & 8.50 & 0.00 \\
OCR (4) & 36.11 & 26.66 & 6.90 & 7.57 & 0.00 \\
Ref (5) & 8.58 & 7.03 & 0.21 & 3.00 & 0.10 \\
\midrule
\multicolumn{6}{l}{\textbf{Rehearsal \((1\%)\)}} \\
\midrule
Instruct (2) & --- & --- & --- & --- & --- \\
VQA (3) & 45.52 & 32.23 & 6.59 & 9.36 & 0.00 \\
OCR (4) & 25.78 & 21.39 & 1.07 & 7.41 & 0.00 \\
Ref (5) & 23.49 & 14.75 & 1.42 & 6.87 & 2.25 \\
\midrule
\multicolumn{6}{l}{\textbf{mSGM + Rehearsal \((1\%)\)}} \\
\midrule
Instruct (2) & 0.10 & 0.00 & 0.13 & 0.00 & 0.00 \\
VQA (3) & 42.85 & 29.98 & 13.41 & 8.35 & 0.00 \\
OCR (4) & 40.51 & 29.30 & 7.64 & 7.50 & 0.00 \\
Ref (5) & 39.66 & 28.81 & 13.89 & 7.99 & 4.30 \\
\end{longtable}}

\subsection{Pre-trained LLMs\label{sec:llms-links}}

Here we provide clickable links to download each of the open-source pre-trained LLMs used in this paper:
\begin{itemize}[noitemsep,nolistsep]
    \item \href{https://huggingface.co/microsoft/phi-2}{phi-2-3b}
    \item \href{https://huggingface.co/EleutherAI/pythia-160m-deduped}{pythia-160m}
    \item \href{https://huggingface.co/EleutherAI/pythia-410m-deduped}{pythia-410m}
    \item \href{https://huggingface.co/EleutherAI/pythia-1b-deduped}{pythia-1b}
    \item \href{https://huggingface.co/EleutherAI/pythia-1.4b-deduped}{pythia-1p4b}
    \item \href{https://huggingface.co/lambdalabs/pythia-1.4b-deduped-synthetic-instruct}{pythia-1p4b-instruct}
    \item \href{https://huggingface.co/EleutherAI/pythia-2.8b-deduped}{pythia-2p8b}
    \item \href{https://huggingface.co/meta-llama/Llama-2-7b-hf}{llama2-7b-pure}
    \item \href{https://huggingface.co/lmsys/vicuna-7b-v1.5}{vicuna-v15-7b}
\end{itemize}

\subsection{Artifact Use}

We ensure that any artifacts (such as datasets, software, models, code, or other supplementary materials) associated with our paper are used in a manner that aligns with their original purpose and the guidelines set forth by the creators. In particular, our artifacts are the models and code we build our experiments on top of, which we have listed above in Sections~\ref{sec:llms-links}.

\end{document}